\documentclass[runningheads]{llncs}
\usepackage[T1]{fontenc}
\usepackage{graphicx}
\usepackage{color}

\usepackage{multirow}
\newcommand{\bftab}{\fontseries{b}\selectfont}
\usepackage{amsmath}
\usepackage{subcaption}

\DeclareMathOperator{\sign}{sign}

\newcommand{\figref}[1]{\figurename~\ref{#1}}
\newcommand{\tabref}[1]{\tablename~\ref{#1}}

\usepackage[hidelinks]{hyperref}

\begin{document}

\title{Deep Reinforcement Learning Based System for Intraoperative Hyperspectral Video Autofocusing}
\titlerunning{DRL Based System for Intraoperative Hyperspectral Video Autofocusing}

\author{
Charlie Budd\inst{1}\thanks{CONTACT Charlie Budd. Email: charles.budd@kcl.ac.uk}
\and Jianrong Qiu\inst{2}
\and Oscar MacCormac\inst{1,3}
\and Martin Huber\inst{1}
\and Christopher Mower\inst{1}
\and Mirek Janatka\inst{1,4}
\and Théo Trotouin\inst{1,4}
\and Jonathan Shapey\inst{1,4}
\and Mads S. Bergholt\inst{2}
\and Tom Vercauteren\inst{1,4}
}
\authorrunning{C. Budd et al.}
\institute{
King’s College London, Biomedical Engineering \& Imaging Science, London
\and King’s College London, School of Craniofacial and Regenerative Biology, London
\and King’s College Hospitals, Department of Neurosurgery, Denmark Hill, London
\and Hypervision Surgical Limited, 1st Floor 85 Great Portland Street, London
}
\maketitle

\begin{abstract}
Hyperspectral imaging (HSI) captures a greater level of spectral detail than traditional optical imaging, making it a potentially valuable intraoperative tool when precise tissue differentiation is essential.
Hardware limitations of current optical systems used for handheld real-time video HSI result in a limited focal depth, thereby posing usability issues for integration of the technology into the operating room.
This work integrates a focus-tunable liquid lens into a video HSI exoscope, and proposes novel video autofocusing methods based on deep reinforcement learning.
A first-of-its-kind robotic focal-time scan was performed to create a realistic and reproducible testing dataset.
We benchmarked our proposed autofocus algorithm against traditional policies,
and found our novel approach to perform significantly ($p<0.05$) better than traditional techniques ($0.070\pm.098$ mean absolute focal error compared to $0.146\pm.148$).
In addition, we performed a blinded usability trial by having two neurosurgeons compare the system with different autofocus policies, and found our novel approach to be the most favourable, making our system a desirable addition for intraoperative HSI.
\end{abstract}

\section{Introduction}

\subsection{Background}
Traditional optical imaging samples the visual spectrum in three diffuse spectral bands (RGB), while hyperspectral imaging (HSI) provides much more detailed spectral information.
This information is potentially valuable for making intraoperative decisions, particularly in cases where tissue differentiation is critical but challenging to perform using traditional visualisation techniques.
In the case of brain tumour excision, fluorescence-guided resection is commonly used to minimize damage to healthy tissue~\cite{Bogaards2004} but is limited to high-grade gliomas, and results in added cost and workflow disruptions.
Thanks to a more detailed definition between tissue types~\cite{Fabelo2013}, HSI is seen as a promising alternative with wider applicability and smoother integration into the workflow.

\begin{figure}
\centering
\begin{subfigure}{0.36\textwidth}
\includegraphics[width=\textwidth,trim={5cm 11cm 10cm 20cm},clip]{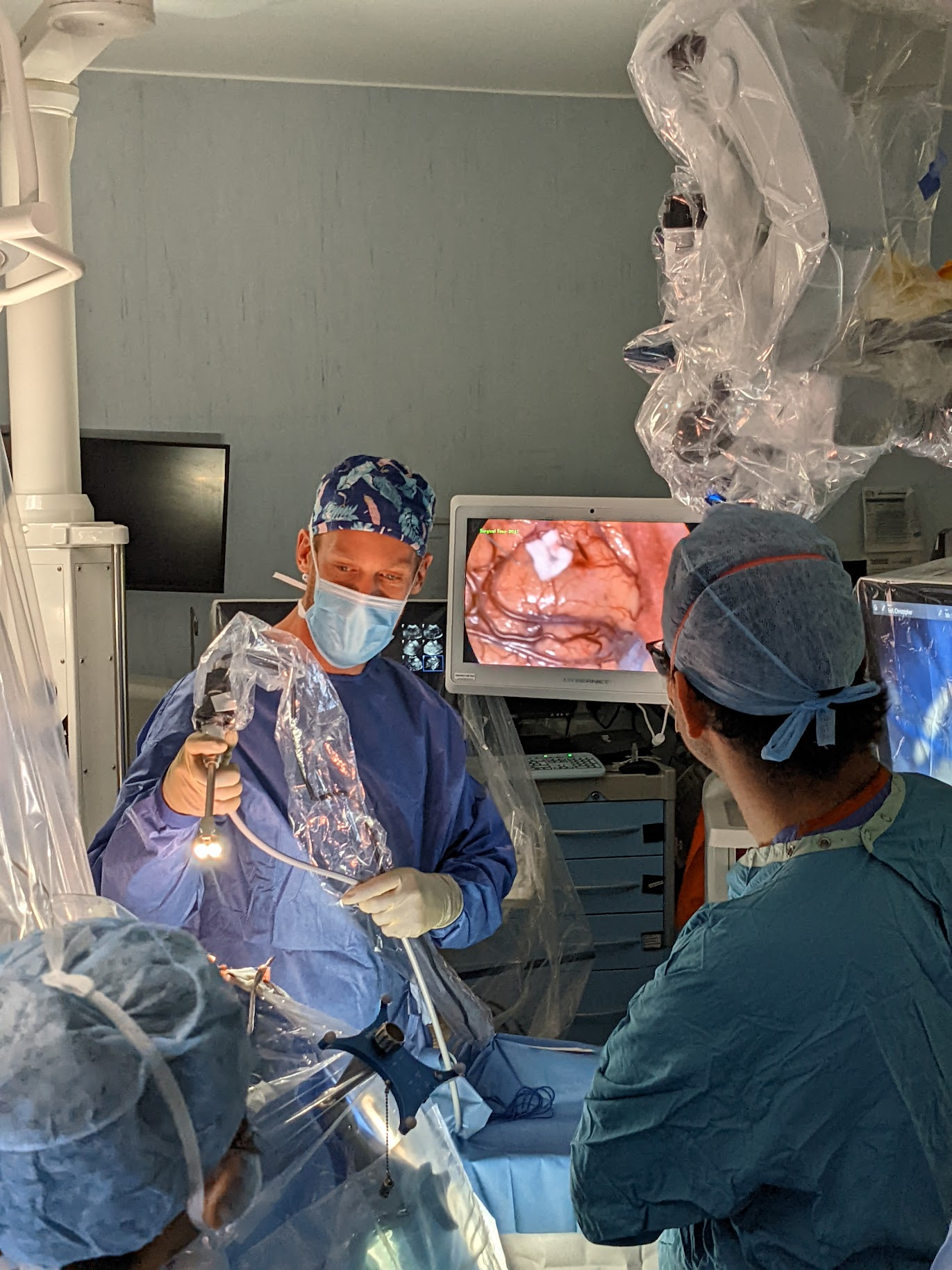}
\end{subfigure}
\begin{subfigure}{0.625\textwidth}
\includegraphics[width=\textwidth,trim={0cm 0cm 5cm 0cm},clip]{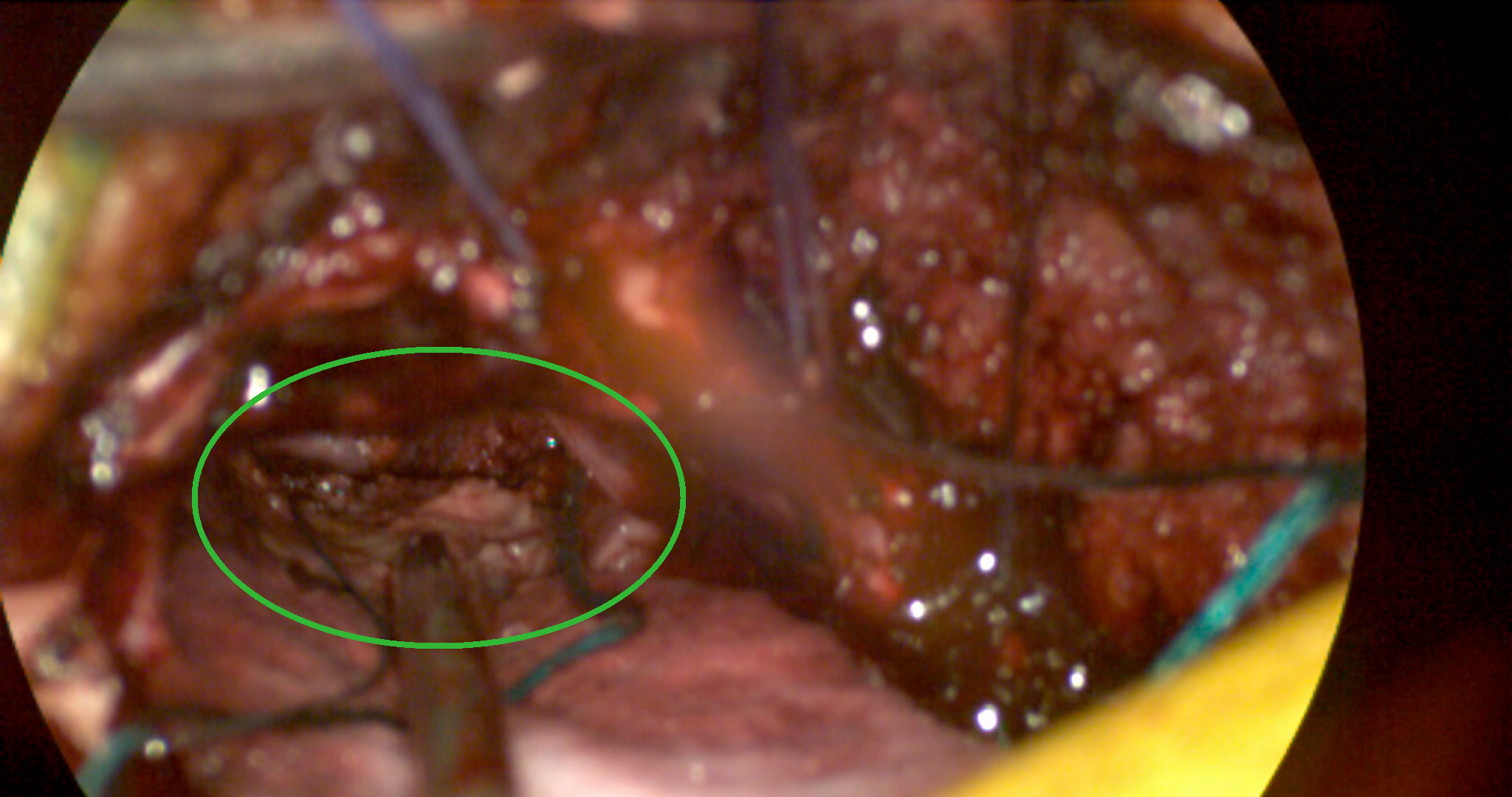}
\end{subfigure}
\caption{
Left) Existing fixed-focus HSI system being used during neurosurgery
in an ethically approved study.
Right) RGB reconstruction of an image taken with the fixed-focus HSI system following a craniotomy. The focus has been manually adjusted for the cavity visible through the craniotomy (circled).
}
\label{realimage}
\end{figure}

While HSI has been integrated into surgical microscope systems~\cite{Pichette2016}, it is suggested that handheld systems are better suited to translational research~\cite{Ebner2021}.
Such handheld systems consist of an exoscope coupled to a draped optical stack, as shown in~\figref{realimage}.
The optics in the exoscope typically result in a short focal depth, making manual focusing tricky, particularly as the tuning must be performed through the drape.
As such, these systems are commonly left at a fixed focal power and the surgeon must keep the working distance fixed to keep the subject in focus.
Furthermore, the narrow spectral bands of HSI sensors reduce the amount of light collected~\cite{Shapey2019}.
To avoid increasing exposure time, a large aperture size is needed, at a cost of further reducing focal depth.
This exacerbates the focusing issues, making current real-time handheld HSI imaging systems particularly challenging to focus, posing significant usability issues.
\figref{realimage} highlights the limited focal depth of our system, and shows a typical target that the surgeon must manually bring into focus during surgery.

The issue of reduced focal depth in real-time HSI systems could be mitigated by the introduction of a video autofocus system.
Autofocus methods are divided into active methods, which use transmission to probe the scene, and passive methods, which rely only on incoming light.
Passive methods are further split into phase-based, which require specialised hardware, and contrast-based, which compare images captured at different focal powers. Our investigation focuses on contrast-based methods, which require minimal hardware development.

\subsection{Related Autofocusing Works}
While autofocusing systems are prevalent in consumer device, the scientific literature is sparse, especially for dynamic video autofocusing. Many publications in the field are concerned with benchtop microscope autofocus systems \cite{Jia2022,Yu2018,Tonislav2020}.
This environment is conducive to autofocus as the scene is typically static with a single focal plane across the whole image.
Additionally, the focus can be adjusted easily by moving the stage vertically.
\cite{Jia2022} take a traditional approach, making use of a Laplacian focal metric combined with a modified hill-climber optimisation scheme.
\cite{Yu2018} input a stack of sequential images to a 3d convolutional neural network (CNN) trained as a deep reinforcement agent trained to output changes in stage height.
\cite{Tonislav2020} train a CNN to regress the optimal focal power from just two images taken at different focal powers.
Beyond benchtop microscopy, \cite{Herrmann2020} also use a CNN to directly regress optimal focal powers, this time from varying number of samples from the full focal stacks.
\cite{Anikina2021} take a novel approach by using pre-trained object detection models to generate latent vector representations of images and using these as inputs to a deep reinforcement agent.
\cite{Wang2021} train two CNNs, one to regress focal steps from a single image, the other to determine if the current image is in focus.

\subsection{Contributions}
This work aims to improve intraoperative handheld HSI systems by alleviating one of their main usability drawbacks, that of shortened focal depth.
We introduce an autofocus system to an existing handheld intraoperative real-time HSI system~\cite{Ebner2021}.
The focus adjustments are handled by a focus tunable liquid lens which is integrated into the setup.
We propose autofocusing policies based on deep reinforcement learning and compare these to traditional heuristic approaches.
Our final model is similar to that presented in \cite{Yu2018} but differs in its use of a weight shared image encoder, software simulated defocusing for training data, and small input patch size.
In addition, our method is designed and trained to handle dynamic environments, something entirely missing in the literature.
We performed a robotic focal-time scan to create a reproducible testing benchmark and allow quantitative comparison of autofocus policies.
Finally, we demonstrate the utility of our approach in a blinded user study involving two neurosurgeons.

\section{Materials and Methods}

\subsection{Optical System}
\begin{figure}
\centering
\includegraphics[width=1.0\textwidth,trim={0 12.7cm 0 0.8cm},clip]{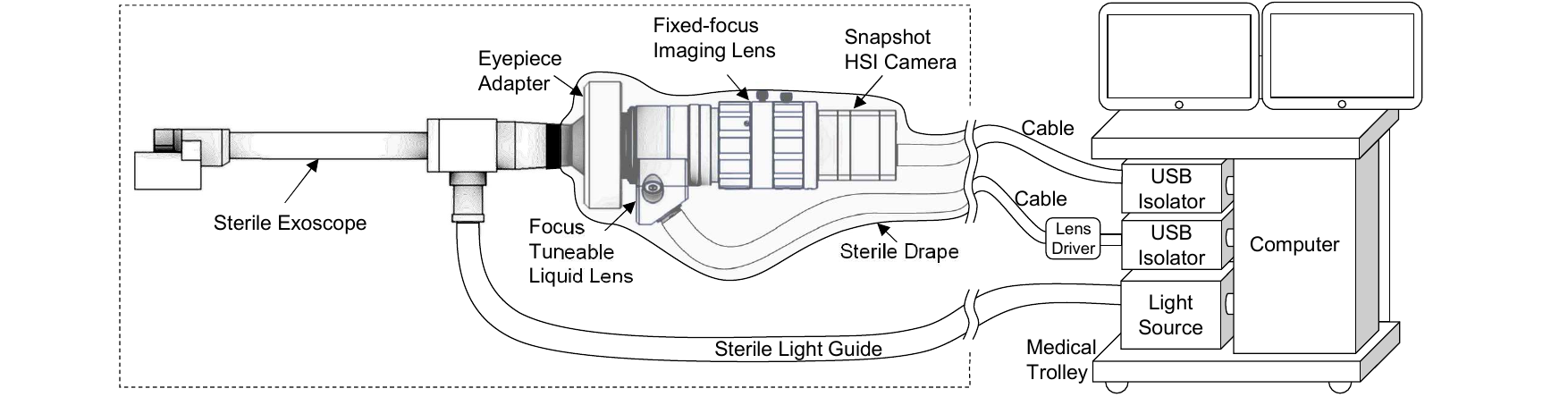}
\caption{Schematic diagram of our intraoperative video HSI system with focus-tunable liquid lens, allowing electrically controllable focal length. The handheld portion of the system is shown in the dashed line box.}
\label{optical_setup}
\end{figure}
Our intraoperative HSI system, shown in \figref{optical_setup}, builds on our existing system~\cite{Ebner2021} by integrating an Optotune EL-10-30-Ci focus-tunable liquid lens to allow electrical control of the focal length.
The hyperspectral camera is based on an IMEC 2/3'' snapshot mosaic CMV2K-SSM4X4-VIS sensor, which acquires 16 spectral bands in a 4$\times$4 mosaic between the spectral range of 460 nm and 600 nm.
With a sensor resolution of 2048$\times$1088 pixels, hyperspectral data is acquired with a spatial resolution of 512$\times$272 pixels per spectral band.
Video-rate imaging of snapshot data is achieved with a speed of up to 50 FPS depending on acquisition parameters.

\subsection{Datasets}

\subsubsection{Software Simulated Focal-Time Scans}
We define a focal-time scan as a time series of focal stacks, with a focal stack being a single image captured at multiple focal lengths.
In order to assemble a large and diverse focal-time scan dataset, we choose to simulate focal-time scans using existing in-focus video data.
To ensure the resulting focal-time scan features diverse camera motion, we implement a smooth random walk to step a cropping rectangle across the video after each frame.
This also allows for the construction of plausible focal-time scans from single images, although features such as dynamic subjects or imaging noise will be missing.
In order to simulate defocus, we implement another random walk to simulate a dynamic optimal focal power.
When an agent is interacting with the simulated scan, a Gaussian filter is used to approximate focal blurring with $\sigma=\sigma_0|f^*-f|$ where $f$ and $f^*$ are the current and optimal focal powers and $\sigma_0$ is chosen randomly from the range $2$--$8$ for each scan.
We use this technique to create a training and testing dataset consisting of 1000 and 200 simulated focal-time scans based on 200 10-second video clips sampled from Cholec80~\cite{Twinanda2016}, a popular endoscopic dataset.
In addition, we created simulated focal-time scans from 200 in focus images taken of a brain phantom with our HSI system. 
These act as a validation dataset to help prevent over fitting and aid generalisation.
While Gaussian blur is a reasonable approximation, we note that more rigours methods exist to simulate defocus blur that may produce better simulated data~\cite{Liu2021}.

\subsubsection{Robotic Focal-Time Scan}
\begin{figure}[t]
\centering
\begin{subfigure}{0.338\textwidth}
\includegraphics[width=\textwidth,trim={0 8cm 0 8cm},clip]{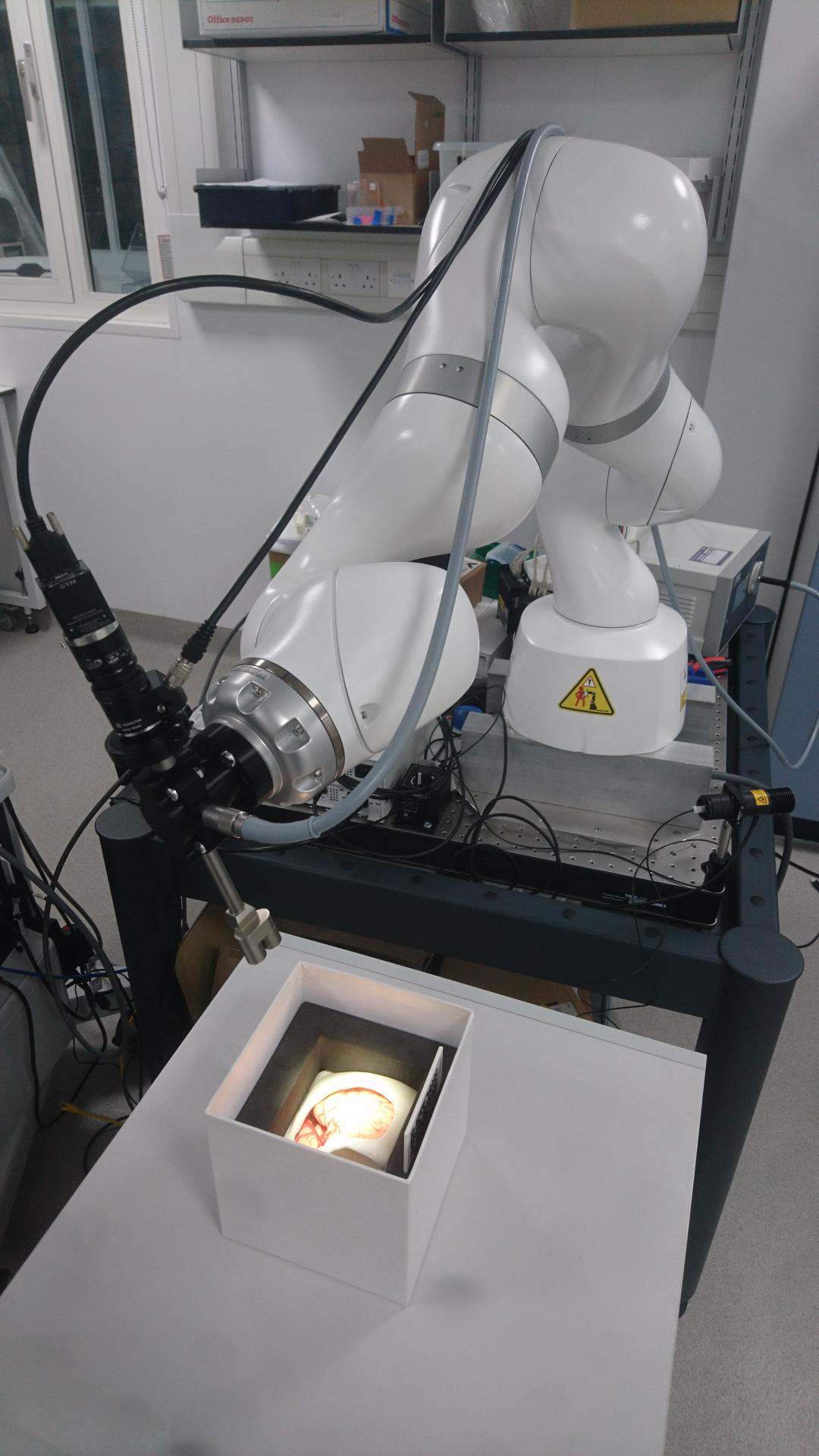}
\end{subfigure}
\begin{subfigure}{0.65\textwidth}
\includegraphics[width=\textwidth,trim={1.5cm 1.5cm 1.5cm 0cm},clip]{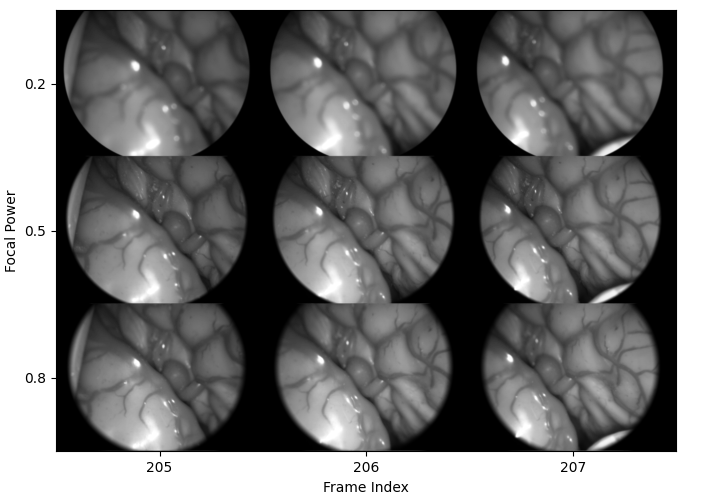}
\end{subfigure}
\caption{
Left) Robotic arm holding our optical system imaging a brain phantom.
Right) Sample from our robotic focal-time scan, with the columns representing sequential focal stacks sampled at focal powers of 0.2, 0.5 and 0.8 (top to bottom).
For low focal powers, the focal plane is behind the phantom (upper row). As the focal power increases, the focal plane intersects with the fissure (middle row), and then with the area surrounding the fissure (bottom row).
}
\label{robotscan}
\end{figure}

As a testing dataset similar to our intended use case, we chose to approximate a real focal-time scan by controlling conditions during capture of the individual focal stacks.
Our optical system was fixed to a robotic arm, which was then used in a compliant control mode to record a natural hand-guided trajectory whilst imaging a brain phantom.
The motion was performed to try to emulate typical usage during a surgery, whilst also trying to cover the range of plausible working distances.
The focal range of the liquid lens is discretised into a set of focal powers, and the recorded trajectory is discretised into a sequence of 1184 poses. 
For each discrete pose, the robotic arm is fixed, and an image captured for each focal power.
We randomise the order of the focal powers to reduce systematic bias caused by the response of the liquid lens.
Auto-exposure was implemented in order to ensure good exposure across all working distances.
To ensure consistency within a given focal stack, auto-exposure was only stepped in-between discrete poses.
The robotic arm holding our optical system and a sample of the resulting focal-time scan can be seen in~\figref{robotscan}.
The optimal focus for all focal stacks was computed via global search of a traditional focal metric (mean gradient magnitude) as detailed below.
This was then validated visually and corrected where appropriate.

\subsubsection{Integration and Usability Trial}
To ensure the validity of our quantitative evaluation, and to get feedback on the system in general, a blinded trial was set up with two practising neurosurgeons.
A set was made containing two repeats of three selected autofocus policies.
This set was then shuffled, and the surgeons remained blinded to the autofocus policy until after the trial.
Each surgeon used our optical system to inspect a brain phantom with each policy in the set.
The surgeon was made aware when the policy was changed and prompted to make comments throughout the trial, which were recorded.

\subsection{Autofocus Policies}
As seen in \figref{realimage}, the area of surgical interest can make up a rather small amount of the overall image, as such, we limit ourselves to a patch size of just $32\times32$ pixels.
The positioning of the patch could be dictated by a second algorithm or user input, but this is outside the scope of this work.
Here, we simply position the patch at the centre of the circular content area, which is detected using the method presented in \cite{Budd2023}.
All of our autofocus policies deal with the grayscale reconstruction of the HSI images.
Throughout this work, we further deal with a normalised focal power range (0--1).

\subsubsection{Traditional Approach}
We implement two traditional autofocus policy based on different focal metrics combined with a simple hill-climber optimisation policy.
We choose mean gradient magnitude (MGM) and mean local ratio (MLR). Two focal metrics which are conceptually simple but competitive~\cite{Herrmann2020} and implemented in quite different ways.
They are defined as
\begin{align}
    \phi^{\textrm{MGM}}(I) &= \frac{1}{n}\sum_{p}{\sqrt{{I_x}^2(p) + {I_y}^2(p)}}\\
    \phi^{\textrm{MLR}}(I) &= \frac{1}{n}\sum_{p}{\max{\left(
        \frac{G_\sigma(I)(p) + 1}{I(p) + 1},
        \frac{I(p) + 1}{G_\sigma(I)(p) + 1}
    \right)}}
\end{align}
where $p$ is the set of all pixels in the image, $I_x$ and $I_y$ are defined as the $x$ and $y$ responses of a Sobel filter, and $G_\sigma$ is a Gaussian blur.
The kernel size is chosen as $\sigma=4$ for all our experiments.
Our hill-climber optimisation policy $O^{\textrm{HC}}$
sets the focal power $f$ at time $t+1$ based on information at time $t$ and
is defined as
\begin{equation}
    f_{t+1} = O^{\textrm{HC}}(\phi_t, f_t, \phi_{t-1}) = 
    \begin{cases}
        f_t + d_{prev}h,& \text{if}\: 0 < f_t < 1 \: \text{and} \: \phi_t > \phi_{t-1}\\
        f_t - d_{prev}h,& \text{otherwise}
    \end{cases}
\end{equation}
where $d_{prev} = \sign(f_t - f_{t-1})$ is the direction of the previous step and $h$ is a step size which we set to $h=0.05$ for all our experiments.
We note that our definition is different from standard hill-climber. 
A normal hill-climber will repeat a step while the focal metric is increasing, and either stop or change direction with a smaller step size when the focal metric decreases, but this does not translate to a continuous and dynamic environment.

\subsubsection{Learned Optimisation Policy}
Due to our dynamic environment, it seems likely that considering a sequence of the $N$ last focal metrics, rather than the last two, would help to build a strong optimisation policy.
However, as $N$ increases, it quickly becomes unclear how to incorporate this information effectively.
It is likely that a learning based solution would uncover a better strategy than heuristic approaches. While regression based approcahes may work, reinforcement learning provides a natural framework for this problem by allowing the policy to model the trade-off between maximisation and exploration.
By modelling the autofocus task as a Markov process, we can define a Q-function $Q(s,a)$ which maps state-action pairs to expected future rewards.
We define our state, actions, and reward function as
\begin{align*}
    s_t&=\{\phi_t, f_t,...,\phi_{t-(N-1)},f_{t-(N-1)}\} \\
    A&=\{-h, 0, +h\}\\
    r_t&=-|f^*_t - f_t|
\end{align*}
where $f^*_t$ is the optimal focal power at $t$
which can only be known in controlled environment.
As before, we take $h=0.05$.
Our learned optimisation policy $O_{RL}$ can then be defined as
\begin{equation}
    f_{t+1} = O^{\textrm{RL}}(f_{t}, s_t) = f_{t} + \max_{a}Q(s_t, a)
\end{equation}
To model $Q(s,a)$, we use an MLP consisting of 2 hidden layers of 256 ReLUs each and a third layer with 3 outputs corresponding to the 3 possible actions.
The MLP takes as input the state vector $s$ containing the $N$ most recent focal metrics and focal powers, we take $N=8$ for all our experiments. 
To train the model, we use Deep Q Learning following the recommendations set out by the DQN method~\cite{Mnih2013} to improve training stability.
We use an experience memory with size $2.5\times10^6$, and an $\epsilon$-greedy exploration policy where $\epsilon$ exponentially decays from $1.0$ to $0.1$ over the first $2\times10^6$ experiences.
Our target model is updated with exponential moving average (EMA) weight updates with a $\beta=0.005$, and we use $\gamma=0.99$ in our Bellman equation.
Finally, we use a smoothed L1 loss function and optimise with RMSProp with learning rate $1\times10^{-5}$ and momentum $0.95$.
We trained on our software simulated focal-time scans created from real endoscopy videos and validated against our simulated focal-time scans created with HSI images taken with our optical system mounted on a robotic arm.

\subsubsection{End-to-End Model}
In addition to learning the optimisation policy, we can also learn the focal metric.
By learning the two together, we are no longer constrained to a scalar metric and can instead learn a latent vector encoding of the image patches. 
To do this, we construct a CNN consisting of 4 convolutions with 8 filters each and a stride of 2, outputting a vector of 8 logits for our patch size of $32\times32$.
The CNN is run on each of the $N$ most recent image patches as a batch during training, but only the most recent during inference, with the previous encodings stored between steps.
The encodings are concatenated with the $N$ most recent focal powers and fed into an MLP.
The MLP and training procedure are the same as before.

\section{Results}
\begin{figure}
\centering
\includegraphics[width=\textwidth,trim={0 0 0 0.9cm},clip]{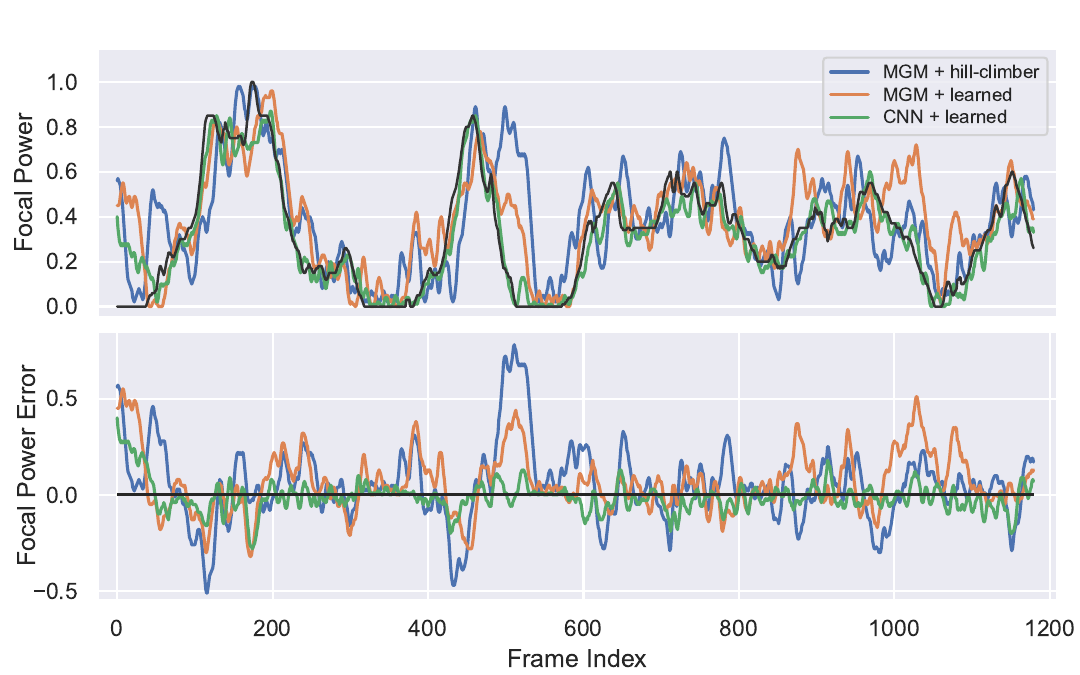}
\caption{Focal path (top) and error in focal power (bottom) for three autofocus policies on the robotic focal-time scan. The optimal focal power is shown in black. All paths have been smoothed with a moving average with a window of 5 frames for visualisation purposes.}
\label{focalpaths}
\end{figure}

We evaluated each autofocus policies on both our simulated focal-time scan test set, and the robotically recorded focal-time scan. The mean focal errors are shown in \tabref{results}.
The scores show an improvement in almost all cases by the introduction of a learned optimiser.
The paths taken and the focal error over time for the robotic focal-time scan for a selection of policies are plotted in \figref{focalpaths}.

During the usability trial, the surgeon participants were positive about all presented policies.
In line with our quantitative results,
the participants both showed preference for the CNN-based policy.
It was thought by both to be smoother and more deliberate in its adjustments, and felt more robust to minor accidental motions inherent to hand-operated system.
One commented that it felt slower to focus but more stable, going on to state that this was desirable behaviour.
All algorithms handled the brain fissure well, this is likely due to the small patch size used, allowing for precise targeting.
Overall, the surgeons were very positive about the integration of autofocus into optical imaging systems.

\begin{table}
\centering
\caption{Mean absolute focal power error (0 to 1), and the percentage of in focus frames (focal power error < 0.1), for different autofocus policies on both the simulated and robotic focal-time scan testing sets.}
\label{results}
\begin{tabular}{|c|c|c|c|c|c|}
\hline
\multicolumn{2}{|c|}{Autofocus policy} & \multicolumn{2}{|c|}{Focal power error (MAE)} & \multicolumn{2}{|c|}{In focus frames}\\
\hline
 Metric & Optimiser & Simulated & Robotic & Simulated & Robotic \\
\hline
n/a & fixed           & 0.236$\pm$.152 & 0.262$\pm$.161 & 19.0\% & 26.4\% \\
MGM & hill-climber    & 0.102$\pm$.138 & 0.146$\pm$.148 & 67.9\% & 46.4\% \\
MLR & hill-climber    & 0.092$\pm$.118 & 0.163$\pm$.168 & 68.2\% & 44.1\% \\
MGM & learned         & 0.085$\pm$.115 & 0.126$\pm$.118 & 70.4\% & 50.8\% \\
MLR & learned         & 0.098$\pm$.120 & 0.156$\pm$.131 & 66.4\% & 39.5\% \\
CNN & learned         & \bftab{0.049$\pm$.072} & \bftab{0.070$\pm$.099} & \bftab{84.9\%} & \bftab{79.1\%} \\
\hline
\end{tabular}
\end{table}

\section{Conclusion}
We have successfully designed a handheld intraoperative HSI imaging system with autofocusing capability.
We developed a novel CNN-based autofocus policy suitable for video data.
In addition, we performed a robotic focal-time scan to evaluate our methods.
Our novel method significantly outperforms a traditional baseline on our robotic focal-time scan, and performs preferably in a usability trial by two neurosurgeons. 
The comments from the usability trial also suggest that the dynamic video autofocusing systems will be well received among surgeons.

\subsubsection{Acknowledgements}
This study/project is funded by the NIHR [NIHR202114].
This work was supported by core funding from the Wellcome/EPSRC [WT203148/Z/16/Z; NS/A000049/1].
This project has received funding from the European Union's Horizon 2020 research and innovation programme under grant agreement No 101016985 (FAROS project).
TV is supported by a Medtronic / RAEng Research Chair [RCSRF1819\textbackslash7\textbackslash34].
For the purpose of open access, the authors have applied a CC BY public copyright licence to any Author Accepted Manuscript version arising from this submission.
TV is a co-founder and shareholder of Hypervision Surgical.

\bibliographystyle{splncs04}

\begin{thebibliography}{10}
\providecommand{\url}[1]{\texttt{#1}}
\providecommand{\urlprefix}{URL }
\providecommand{\doi}[1]{https://doi.org/#1}

\bibitem{Anikina2021}
Anikina, A., Rogov, O.Y., Dylov, D.V.: Dasha: Decentralized autofocusing system
    with hierarchical agents. arXiv preprint arXiv:2108.12842  (2021)

\bibitem{Bogaards2004}
Bogaards, A., Varma, A., Collens, S.P., Lin, A., Giles, A., Yang, V.X., Bilbao,
    J.M., Lilge, L.D., Muller, P.J., Wilson, B.C.: Increased brain tumor
    resection using fluorescence image guidance in a preclinical model. Lasers in
    Surgery and Medicine  \textbf{35}(3),  181--190 (2004).
    \doi{https://doi.org/10.1002/lsm.20088},
    \url{https://onlinelibrary.wiley.com/doi/abs/10.1002/lsm.20088}

\bibitem{Budd2023}
Budd, C., Herrera, L.C.G.P., Huber, M., Ourselin, S., Vercauteren, T.: Rapid
    and robust endoscopic content area estimation: a lean gpu-based pipeline and
    curated benchmark dataset. Computer Methods in Biomechanics and Biomedical
    Engineering: Imaging \& Visualization  \textbf{0}(0),  1--10 (2023).
    \doi{10.1080/21681163.2022.2156393},
    \url{https://doi.org/10.1080/21681163.2022.2156393}

\bibitem{Ebner2021}
Ebner, M., Nabavi, E., Shapey, J., Xie, Y., Liebmann, F., Spirig, J.M., Hoch,
    A., Farshad, M., Saeed, S.R., Bradford, R., Yardley, I., Ourselin, S.,
    Edwards, D., Fuhrnstahl, P., Vercauteren, T.: Intraoperative hyperspectral
    label-free imaging: From system design to first-in-patient translation.
    JOURNAL OF PHYSICS D APPLIED PHYSICS  \textbf{54}(29) (2021).
    \doi{10.1088/1361-6463/abfbf6},
    \url{http://www.scopus.com/inward/record.url?scp=85107008535&partnerID=8YFLogxK},
    10.1088/1361-6463/abfbf6

\bibitem{Fabelo2013}
Halicek, M., Fabelo, H., Ortega, S., Callico, G.M., Fei, B.: In-vivo and
    ex-vivo tissue analysis through hyperspectral imaging techniques: Revealing
    the invisible features of cancer. Cancers  \textbf{11}(6) (2019).
    \doi{10.3390/cancers11060756}, \url{https://www.mdpi.com/2072-6694/11/6/756}

\bibitem{Herrmann2020}
Herrmann, C., Bowen, R.S., Wadhwa, N., Garg, R., He, Q., Barron, J.T., Zabih,
    R.: Learning to autofocus. In: Proceedings of the IEEE/CVF Conference on
    Computer Vision and Pattern Recognition (CVPR) (June 2020)

\bibitem{Tonislav2020}
Ivanov, T., Kumar, A., Sharoukhov, D., Ortega, F., Putman, M.: {DeepFocus: a
    deep learning model for focusing microscope systems}. In: Zelinski, M.E.,
    Taha, T.M., Howe, J., Awwal, A.A.S., Iftekharuddin, K.M. (eds.) Applications
    of Machine Learning 2020. vol. 11511, p. 1151103. International Society for
    Optics and Photonics, SPIE (2020). \doi{10.1117/12.2568990},
    \url{https://doi.org/10.1117/12.2568990}

\bibitem{Jia2022}
Jia, D., Zhang, C., Wu, N., Zhou, J., Guo, Z.: Autofocus algorithm using
    optimized laplace evaluation function and enhanced mountain climbing search
    algorithm. Multimedia Tools and Applications  \textbf{81}(7),  10299--10311
    (2022)

\bibitem{Liu2021}
Liu, Y.Q., Du, X., Shen, H.L., Chen, S.J.: Estimating generalized gaussian blur
    kernels for out-of-focus image deblurring. IEEE Transactions on Circuits and
    Systems for Video Technology  \textbf{31}(3),  829--843 (2021).
    \doi{10.1109/TCSVT.2020.2990623}

\bibitem{Mnih2013}
Mnih, V., Kavukcuoglu, K., Silver, D., Graves, A., Antonoglou, I., Wierstra,
    D., Riedmiller, M.: Playing atari with deep reinforcement learning. arXiv
    preprint arXiv:1312.5602  (2013)

\bibitem{Pichette2016}
Pichette, J., Laurence, A., Angulo, L., Lesage, F., Bouthillier, A., Nguyen,
    D.K., Leblond, F.: {Intraoperative video-rate hemodynamic response assessment
    in human cortex using snapshot hyperspectral optical imaging}. Neurophotonics
    \textbf{3}(4),  045003 (2016). \doi{10.1117/1.NPh.3.4.045003},
    \url{https://doi.org/10.1117/1.NPh.3.4.045003}

\bibitem{Shapey2019}
Shapey, J., Xie, Y., Nabavi, E., Bradford, R., Saeed, S.R., Ourselin, S.,
    Vercauteren, T.: Intraoperative multispectral and hyperspectral label-free
    imaging: A systematic review of in vivo clinical studies. Journal of
    biophotonics  \textbf{12}(9),  e201800455 (2019)

\bibitem{Twinanda2016}
Twinanda, A.P., Shehata, S., Mutter, D., Marescaux, J., de~Mathelin, M., Padoy,
    N.: Endonet: {A} deep architecture for recognition tasks on laparoscopic
    videos. CoRR  \textbf{abs/1602.03012} (2016),
    \url{http://arxiv.org/abs/1602.03012}

\bibitem{Wang2021}
Wang, C., Huang, Q., Cheng, M., Ma, Z., Brady, D.J.: Deep learning for camera
    autofocus. IEEE Transactions on Computational Imaging  \textbf{7},  258--271
    (2021). \doi{10.1109/TCI.2021.3059497}

\bibitem{Yu2018}
Yu, X., Yu, R., Yang, J., Duan, X.: A robotic auto-focus system based on deep
    reinforcement learning. In: 2018 15th International Conference on Control,
    Automation, Robotics and Vision (ICARCV). pp. 204--209. IEEE (2018)

\end{thebibliography}

\end{document}